\documentclass{article} 
\usepackage{collas2023_conference,times}
\usepackage{easyReview}

\usepackage{amsmath,amsfonts,bm}

\def\eqref#1{equation~\ref{#1}}

\def\1{\bm{1}}

\DeclareMathAlphabet{\mathsfit}{\encodingdefault}{\sfdefault}{m}{sl}
\SetMathAlphabet{\mathsfit}{bold}{\encodingdefault}{\sfdefault}{bx}{n}

\usepackage{hyperref}
\hypersetup{
    colorlinks=true,
    linkcolor=red,
    filecolor=magenta,
    urlcolor=blue,
    citecolor=purple,
    pdftitle={Overleaf Example},
    pdfpagemode=FullScreen,
    }

\usepackage{subcaption}
\usepackage{enumitem,amssymb}
\usepackage{adjustbox,booktabs}
\usepackage{tabularx}
\usepackage{wrapfig}
\usepackage{pifont}
\usepackage{color}
\usepackage{array,multirow}
\usepackage{amsmath}
\usepackage[ruled,vlined]{algorithm2e}

\SetCommentSty{mycommfont}
\usepackage{floatrow}
\usepackage{nccmath}

\usepackage{multibib}

\newcommand{\MSE}[1]{\texttt{MSE}}
\newcommand{\GEN}[1]{\texttt{GEN}}
\newcommand{\OJA}[1]{\texttt{OJA}}
\newcommand{\INEL}[1]{\texttt{INEL}}
\newcommand{\SPARSE}[1]{\texttt{SPARSE}}

\newcolumntype{H}{>{\setbox0=\hbox\bgroup}c<{\egroup}@{}}

\newlist{todolist}{itemize}{2}
\setlist[todolist]{label=$\square$}

\definecolor{darkpastelgreen}{rgb}{0.01, 0.75, 0.24}

\newcommand{\vect}[1]{{#1}}

\title{Improving Performance in Continual Learning Tasks using Bio-Inspired Architectures}

\author{%
  Sandeep Madireddy \\
  Argonne National Laboratory\\
  Lemont, IL, USA\\
  \texttt{smadireddy@anl.gov} \\
  \And
  Angel Yanguas-Gil \\
  Argonne National Laboratory \\
  Lemont, IL, USA\\
  \texttt{ayg@anl.gov} \\
  \And 
  Prasanna Balaprakash \\
  Oak Ridge National Laboratory\\
  Oak Ridge, TN, USA \\
  \texttt{pbalapra@ornl.gov} \\
}

\preprintcopy 

\begin{document}

\maketitle

\begin{abstract}

The ability to learn continuously from an incoming data stream without catastrophic
forgetting is critical to designing intelligent systems. Many approaches to continual 
learning rely on stochastic gradient descent and its variants that employ global error
updates, and hence need to adopt strategies such as memory buffers or replay to
circumvent its stability, greed, and short-term memory limitations.
To address this limitation, we have developed a biologically inspired lightweight
neural network architecture that incorporates synaptic plasticity mechanisms
and neuromodulation and hence learns through local error signals to enable
online continual learning without stochastic gradient descent. 

Our approach leads to superior online continual learning performance on Split-MNIST,
Split-CIFAR-10, and Split-CIFAR-100 datasets compared to other memory-constrained 
learning approaches and matches that of the state-of-the-art memory-intensive replay-based 
approaches.
We further demonstrate the effectiveness of our approach by integrating key design
concepts into other backpropagation-based continual learning algorithms, 
significantly improving their accuracy. Our results provide compelling evidence for
the importance of incorporating biological principles into machine learning models
and offer insights into how we can leverage them to design more efficient and robust
systems for online continual learning.

\end{abstract}

\section{Introduction}

Online continual learning addresses the scenario where a system has to learn and process
data that are continuously streamed, often without restrictions in terms of the distribution of data within
and across tasks and without clearly identified task boundaries~\cite{mai2021online,chen2020mitigating,aljundi2019online}. 
Online continual learning algorithms seek to mitigate catastrophic forgetting at both the data-instance and task level~\cite{chen2020mitigating}.
In some cases, however, such as on-chip learning at the edge, additional considerations such as resource limitations in the hardware, 
data privacy, or data security are also important for online continual learning. 

A key challenge of online continual learning is that it runs counter to the optimal conditions
required for optimization using stochastic gradient descent (SGD)~\cite{parisi2019continual},
which struggles with non-stationary data streams~\cite{,Lindsey_NEURIPS2020}.
On the contrary, biological systems excel at online continual learning. Inspired by the structure
and functionality of the mammal brain, several approaches have adopted replay
strategies to counteract catastrophic forgetting during non-stationary tasks.
However, online continual learning is present even in simpler species, such as invertebrates,
lacking the basic structures required to implement any type of replay. This situation suggests
that the use of local learning rules, combined with other structural features in
the brain of invertebrates, should be enough to achieve online continual learning without needing to keep a memory buffer of past experiences.

Most continual learning approaches in the literature~\cite{hadsell2020embracing,delange2021continual} 
 have been designed for offline continual learning scenarios where non-stationarity is 
assumed just between tasks while the data within tasks are assumed to be i.i.d and multiple passes
are made through them.
Some recent work~\cite{aljundi2019gradient,aljundi2019online,chaudhry2018efficient,chen2020mitigating,caccianew} 
considered online continual learning; however, the majority of the works (see~\cite{chen2020mitigating}) adopt a
task-incremental learning setting where task labels are explicitly provided at
test time. Only a handful of works consider the class-incremental learning scenario~\cite{hsu2018re} that
does not have access to task labels, and is therefore much harder.
To handle this scenario, most of the approaches (which employ SGD and global gradient updates) resort to expensive memory
buffers~\cite{hsu2018re,buzzega2020dark,caccianew} and auxiliary data~\cite{zhang2020class} to
mitigate catastrophic forgetting in class-incremental learning.  

In this work, we introduce the neuromodulated neural architecture (NNA), a biologically inspired architecture
that exhibits excellent
performance in various continual learning tasks. First, we mimic the heterogeneous
plasticity of the insect brain by decoupling our networks in a feature extraction and
a learning component. Second, we use local synaptic plasticity rules to carry out
supervised learning on data streams, where labels are passed as modulatory signals 
akin to what happens in ternary synapses. Third, we introduce novel local learning rules that
mimic the transition from short-term to long-term memory that takes place in individual synapses.

Using this architecture, we demonstrate accuracies (with Split-MNIST, Split-CIFAR-10, and Split-CIFAR-100 datasets) that outperform the memory-free approaches and match the approaches that employ memory buffers, with a shallow network and without memory replay, in both task-incremental and class-incremental settings~\cite{hsu2018re}. 
We also demonstrate that salient design principles from the NNA approach that are effective with mitigate catastrophic forgetting, i.e. the biologically inspired neural architecture and loss formulation, can also be integrated into existing continual learning algorithms and improve their online continual learning performance.

\section{Related Work}

\noindent \textbf{Continual learning:} Several continual learning approaches have been presented in the literature to circumvent catastrophic forgetting that can occur due to the non-stationarity in the data brought about by the task switching. These approaches can loosely be categorized as (1) memory buffer and replay based, which store samples from previously observed tasks and replay them while learning a new task;  (2) regularization based, which impose constraints to boost knowledge retainment; (3) metalearning based, which use a series of tasks to learn a common parameter configuration that is easily adaptable for new tasks; and (4) parameter memory-based, which incorporate the task-specific memories directly onto in the network weights.

The {\it replay}-based approaches maintain an exemplar data buffer of samples from previous tasks. Trade-offs have to be made on the size of this buffer and the bias introduced by the use of data from just the most recent task. Notable approaches in this category include bioinspired dual-memory architecture~\cite{parisi2019continual} and deep generative replay~\cite{shin2017continual}, which involves a cooperative dual model architecture framework inspired by the hippocampus, which retains past knowledge by the concurrent replay of generated pseudo data. Gradient episodic memory~\cite{lopez2017gradient}, generative replay with feedback connections~\cite{van2018generative}, and gradient-based sampling~\cite{aljundi2019gradient} are other examples of replay-based approaches. 

The {\it regularization}-based approach includes algorithms such as elastic weight consolidation (EWC)~\cite{kirkpatrick2017overcoming}, which computes synaptic importance using a Fisher importance matrix-based regularization; synaptic intelligence~\cite{zenke2017continual}, whose regularization penalty is similar to EWC but is computed online at per-synapse level;  and leaefning without forgetting~\cite{li2017learning}, which applies a distillation loss on the attention-enabled deep networks seeking to minimize task overlap.

The {\it metalearning}-based continual learning consists of an inner loop that learns parameters in a prediction network and an outer loop that learns the parameters in a representation learning network. These loops are updated by using different strategies and data splits at meta-training and meta-testing. Recent approaches in this category include online metalearning (OML)~\cite{javed2019meta}, neuromodulated metalearning algorithm (ANML) ~\cite{beaulieu2020learning}, and incremental task-agnostic metalearning (iTAML)~\cite{rajasegaran2020itaml}. In terms of memory, a naive metalearning approach will maintain full task history to update the parameters for each traversal through the loop when a new task is observed. OML and ANML algorithms keep all the tasks in memory but randomly choose the task sequences for meta-training (not just from task history) and meta-testing updates. iTAML, on the other hand, uses data from the current task and a exemplar data (memory buffer) from the previously seen tasks to metalearn.
The {\it parameter memory}-based approaches either have fixed architecture and dedicate different subsets of model parameters for different tasks with task-specific masks~\cite{serra2018overcoming} or dynamically grow the network, with a new branch for each task~\cite{aljundi2017expert}.

\noindent \textbf{Online continual learning:} Most of these approaches however, have been designed for the traditional continual learning scenarios where non-stationarity is assumed between tasks while the data within tasks are assumed to be i.i.d and multiple passes are made through them. Recent works, such as~\cite{aljundi2019gradient,aljundi2019online,chaudhry2018efficient,chen2020mitigating}, started to look at the more challenging online continual learning scenario where the data is seen only once. In this scenario, however, the majority of the works~\cite{chen2020mitigating} have looked at the task-incremental learning setting where  the task labels are explicitly provided at test time. Only a few works~\cite{aljundi2019gradient,aljundi2019online,Lee2020A} have employed the class-incremental learning scenario, which does not have access to task labels and hence is much harder. The restriction of learning with only a single pass over the data makes this even harder. To handle this scenario, most of the approaches resort to memory buffers~\cite{hsu2018re,buzzega2020dark,caccianew} and auxiliary data~\cite{zhang2020class}. 

\noindent \textbf{Synaptic Plasticity and Neuromodulation:} Synaptic plasticity mechanisms and neuromodulation have been successfully adopted in supervised learning. For example,  ~\cite{Miconi2018} adopts Hebbian plasticity for designing plastic neural networks capable of metalearning. Another work~\cite{miconi2020backpropamine} extends the Hebbian plasticity to incorporate neuromodulation for similar tasks, \cite{daram2020exploring} adopts neuromodulated plasticity for one-shot image classification, and \cite{vecoven2020introducing} adopts it for reinforcement learning. Recently ~\cite{thangarasa2020enabling} adopts Hebbian synaptic consolidation combined with regularization-based approaches for task-incremental learning on simple MNIST-based benchmarks. 

\section{Model}

\subsection{Neuromodulated Neural Architecture}

\begin{figure}
  \centering
  \includegraphics[width=10cm]{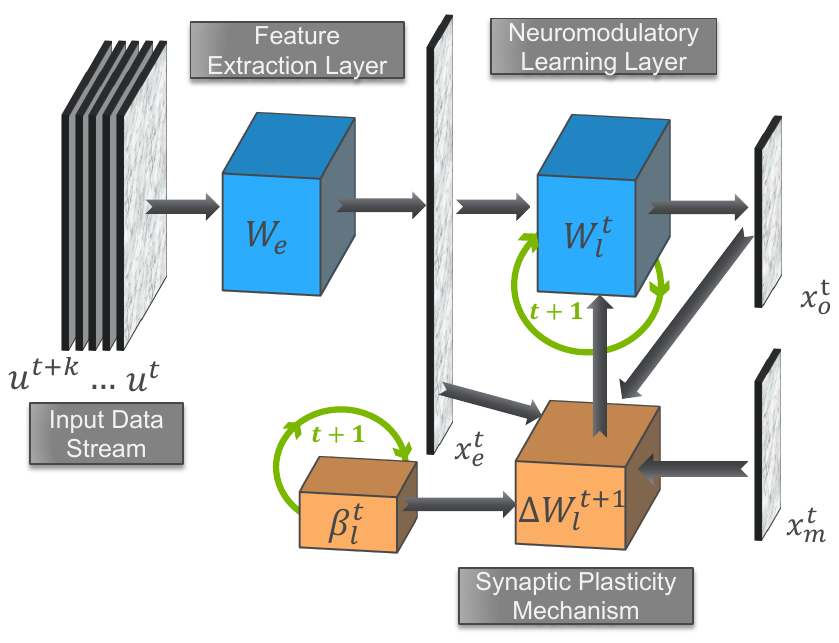}
  \caption{Multilayer neuromodulated architecture: (a) feature extraction, (b) neuromodulated learning layers; enabling synaptic plasticity mechanisms through local learning rules.} 
\label{fig:architecture}
\end{figure}

Our neuromodulated neural architecture (NNA) is elucidated in Fig. \ref{fig:architecture}. Our architecture comprises two components, the feed-forward map $\mathcal{F}$ and the hidden states $S^t$, such that

\begin{equation}
\vect{x}_o^t  =   \mathcal{F}\left(\vect{u}^t, \vect{x}_m^t ;
S^t\right),
\end{equation}
where $\vect{u}^t = \{ x^t, y^t\}$, is the input, $\vect{x}^t_m$ is the modulatory signal evaluated using the ground truth label $\vect{y}^t$, and $\vect{x}_o^t$ is label prediction.
The evolution of the hidden state can be described as
\begin{equation}
S^{t+1} =  \mathcal{G}\left( S^t,\vect{u}^t, \vect{x}_m^t,
\vect{x}_o^t\right),
\end{equation}
where $\mathcal{G}$ is a function that models the evolution of the internal state. A key difference between our formulation of online learning and that of conventional stochastic gradient descent is that the evolution of $S^t$ is computed in real time based solely on the most recent inputs.

We propose a simple architecture that breaks down the input-output map $\mathcal{F}\left(\vect{u}^t, \vect{x}_m^t ; S^t\right)$) into two components: a feature extraction layer~$\mathcal{F}_e$ that transforms input into an internal representation $\vect{x}_e^t  = \mathcal{F}_e(\vect{u}^t, W_e)$ and a neuromodulated learning layer~$\mathcal{F}_l$
that transforms the internal representation into outputs, $\vect{x}_o^t =  \mathcal{F}_l(\vect{x}_e^t, W_l^t)$.
Here, $W_e$ is a projection matrix, while $W_l$ are plastic synaptic weights.

The internal state $S^t$ comprises the synaptic plasticity mechanism $\Delta~W_l^{t}$ and a hyperparameter
vector $\beta_l^t$ that feeds into the synaptic plasticity rule. Therefore,  we can decompose $\mathcal{G}$ into two functions, $\mathcal{F}_w$ and $\mathcal{F}_\beta$, such that $\mathcal{G} = [ \mathcal{F}_w, \mathcal{F}_\beta].$ The function $\mathcal{F}_w$ refers to the synaptic plasticity mechanism such that 
\begin{equation}
  \text{Synaptic Plasticity Mechanism:} \,\,\,\,\,\,\,\, 
  \Delta W_l^{t+1}  =  \mathcal{F}_w\left(\vect{x}_e^{t}, \vect{x}_o^{t},
        \vect{x}_m^{t}; W_l^{t}, \beta_l^{t}\right),
\end{equation}  
with the update for the weights~$W_l^{t}$ in the neuromodulatory layer described as $W_l^{t+1} = W_l^{t} + \Delta W_l^{t+1}.$ Furthermore, the the evolution of hyperparameters $\beta_l^t$ that constitute the synaptic plasticity mechanism is given by
\begin{equation}
  \text{Hyperparameter Evolution:} \,\,\,\,\,\,\,\, \beta_l^{t+1}  =  \mathcal{F}_\beta\left(\vect{x}_e^{t}, \vect{x}_o^{t}, \vect{x}_m^{t}; W_l^{t}, \beta_l^{t}\right).
\end{equation}

Since training this system does not involve storage of batches of prior
data or configurations, any memory of prior
processes has to be built into
the equations for $W_l^t$ and $\beta_l^t$. 
We note that, in contrast to conventional learning approaches,
there is no separation between architecture and
learning algorithm: the ability to learn is defined
by the choice of $\mathcal{F}_w$, $\mathcal{F}_\beta$, and their hyperparameters. 
To this end, a sample from the hyperparameter configuration space comprising
$\left\{\mathcal{F}_e, \mathcal{F}_l, \mathcal{F}_w, \mathcal{F}_\beta, W_l^{t=1}, \beta_l^{t=1}
\right\}$ uniquely defines an NNA.
In the next sections, we describe the specific components chosen in this
work and present the training mechanism. 
The NNA in Figure \ref{fig:architecture} shows one feature extraction and one
neuromodulated learning layer.

\subsubsection{Feature extraction layer} 
\label{feature extraction}
For single-channel datasets we have considered a sparse layer that follows
the same design pattern found in the cerebellum
and the insect's mushroom body \citep{Litwin_2017,ayg_memristors}. 
The $\alpha$-dimensional input ($\vect{u}^t$) is projected into a much larger $\kappa$-dimensional space ($\vect{x}_e^t$) through the sparse projection matrix $W_e$.
We use dynamic thresholding in the sparse layer
followed by a rectified linear unit to ensure
that only the most salient features are included in the representation, which
we refer to as \emph{activity sparsity}. Putting these choices together defines our feature extraction layer $\mathcal{F}_e(\cdot)$ to be
\begin{equation}
\label{eq:dyn_thresh}
    \mathcal{F}_e: \vect{x_e^t} = \mathrm{ReLU}\left(\mathrm{W_e^t} \vect{u^t} - \mu - k \sigma\right),
\end{equation}
where 
$\mu$ and $\sigma$ are respectively the mean and standard deviation of the
matrix multiplication product
$\mathrm{W_e^t} \vect{u^t}$, and $k$ is a constant controlling the cutoff.
This approach highlights correlations between different input channels and represents the broadest possible prior for feature extraction, since it does not assume any spatial dependence or correlations between inputs. Here $\beta_e = \{\kappa, \mu, k, \sigma \}$ 
are all hyperparameters in the feature extraction layer that can be adjusted to the network.

For experiments involving 3-channel images, we transfer feature extractors from various pretrained models. Here, we have used the wide-ResNet50~\citep{zagoruyko2016wide} model pretrained on ImageNet data~\citep{deng2009imagenet} after experimenting with a wide variety of pretrained model and data combinations. Hence, we compose these feature extractors $F_p^e(\cdot)$ with Eqn~\ref{eq:dyn_thresh} to get the feature extraction layer 
\begin{equation}
\label{eq:dyn_thresh2}
    \mathcal{F}_e: \vect{x_e^t} = \mathrm{ReLU}\left(\mathrm{W_e^t} (F_p^e(\vect{u^t})) - \mu - k \sigma\right).
\end{equation}

\subsubsection{Neuromodulated learning layer} 

The neuromodulated learning layer
comprises one or more all-to-all 
connected layers where online supervised learning occurs. $\mathcal{F}_l(\cdot)$ is chosen to be 
\begin{equation}
\label{eq:F_l}
    \mathcal{F}_l: \vect{x_o^t} = \mathrm{Tanh}\left(\mathrm{ReLU}\left(\mathrm{W_l^t}\vect{x}_e^t\right)\right).
\end{equation}

Synaptic weights ($W_l$) are updated in real time through a series of local learning rules ($\mathcal{F}_w(\cdot)$) that provide (local) gradient updates. 
In this work, we focus specifically
on modulated synaptic plasticity rules for supervised learning that are biologically enabled by ternary synapses, where in addition to the presynaptic ($x_e^t$) and postsynaptic ($x_o^t$) inputs, learning is influenced by a local modulatory signal ($x_m^t$) controlling how learning takes place. In this context, the set of hyperparameters would consist of static parameters $\beta^f=\{\beta_1^f,\beta_2^f,\beta_3^f\}$ and dynamic, $\beta_l^t=\{l_r^t\}$ (e.g., learning rate). In particular, we consider the following four learning rules:

\noindent \textbf{Generalized Hebbian model} (GEN): a modulated version of the covariance rule commonly used in neuroscience,  where the synaptic weight evolution is given by
    $\Delta W_l^{t+1} = l_r^t\vect{x}_m^t(\vect{x}_e^t - \beta_1^f)
    (\vect{x}_o - \beta_2^f).$
This rule is well known to be unstable, and 
 clamping or a regularization mechanism
needs to be included in order to keep the synaptic weights
bounded. The rule is at the core of some recent
the neuromodulation-based approaches \citep{Miconi2018}.

\noindent \textbf{Oja's rule} (OJA): it is a heuristic modification of the basic Hebb's rule
providing a normalization mechanisms through a first-order loss term \citep{Oja1982} given by
        $\Delta W_l^{t+1} = l_r^t 
        \vect{x}_m^t(\vect{x}_e^t\vect{x}_o^t - 
        \beta_1^f {x_o^t}^2 W_l^t).$

\noindent \textbf{Mean square error rule} (MSE): a non-Hebbian rule
based on synaptic
plasticity mechanisms in the mushroom body, which is a key memory and
learning center of the insect brain~\citep{Hige_nonhebbian_2015}.
The key assumptions of this rule
are that learning takes place even in absence of postsynaptic activity and is modulated by the difference between the stimulus and the predicted response. We can model this effect using a mean square error (MSE) expression:
        $\Delta W_l^{t+1} = l_r^t
        (\vect{x}_m^t-\vect{x}_o^t)\vect{x}_e^t .$
This rule therefore establishes a connection between a bioinspired process and a cost function used
in stochastic gradient descent methods.

\noindent \textbf{Inelastic learning rule} (INEL): a metaplasticity rule that augments the MSE rule by introducing memory
effects on the synaptic plasticity.  In order
to achieve long-term potentiation, some synapses in the mushroom body require the recurrence of a specific stimulus.
One way of implementing this dependence with past inputs is by considering a synapse-specific window for plasticity based on how significant a weight $W_{ij}$ (which is the $W_l^t$ in equ. 4, where i, j are the indices corresponding to the dimension of $x_e^t$ and $x_o^t$) with respect to its
default value. One simple way
to codify this behavior is to
use the mean synaptic weight of
a layer as a reference
value for the plasticity window. With this assumption, the learning rate at each synapse $K_{ij}$ can
be expressed as
        $K_{ij}^t = H\left(1-\beta_1^f |W_{ij}^t-\mu|\right)$,
Where $H(\cdot)$ is the Heaviside
function and $\mu$ is the average
synaptic weight of each layer.
The resulting $K_{ij}$ is then fed to an MSE rule.
This rule introduces a simple mechanism for memory consolidation based on the inelasticity of the learning rate with respect to the value of the synaptic weight. Moreover, this consolidation is not irreversible: the evolution of the remaining weights can bring weights from frozen to active by reducing its significance.

For all the rules, we have added other conditions to provide symmetric or positive clamping of the synaptic weights in order to prevent instabilities in the algorithm. Finally, the evolution of the hyperparameters with time is given by  
    $\mathcal{F}_\beta: \beta_l^{t+1} = (\beta_l^{t} + \beta_l^{t=1}*\beta_3^f)/ (1+\beta_3^f) $,
The algorithm to train NNA in an online learning scenario is described in Algorithm~\ref{algo1}.

 \begin{figure}[h!]
    \begin{minipage}{\linewidth}
    \small
    \begin{algorithm}[H]
    	{
        \SetAlgoLined
         $acc=0$\;
         $N_{Iter} = N_{Train} \times N_{epochs}$\;
         \While{ t $<$ $N_{Iter}$}{
          $u,y_u$ = Data[$rand^t$]\;
          $x_m$ = zeros($\delta$)\;
          $x_m[y_u]$  = 1 \;
          \tcc{Feature extraction}
          $x_e = F_e(u)$\;
          \tcc{Neuromodulated learning}
          $x_o = F_l(x_e,W_l^t,x_m)$\;%
          \tcc{Synaptic plasticity mechanism}
          $W_l^{t+1} = W_l^{t}+\Delta W_l^{t+1}$ \;
          \If{argmax($x_o == lab$)}{
          $acc=acc+1$\;
          }
          $acc = acc/N_{Iter}$\;
         }
    
         \caption{Training NNA for online learning}
         \label{algo1}
    }
    \end{algorithm}
    \end{minipage}
 \end{figure}

\section{Methodology: Online Learning Scenarios and Architecture Configuration}

\subsection{Online Learning Scenarios }
We have considered experiments in which the system
is subject to a stream of data and labels, evolving 
its internal configuration during a predetermined number of epochs. 
This constitutes a single episode. 
At the end of the episode, the system is evaluated
against the testing dataset to obtain its accuracy. 
By concatenating multiple episodes involving different tasks and
datasets, we can create a curriculum to evaluate the system's ability to carry
out online continual learning.

We consider three learning modalities.
\textit{(1) Single-Task Online Learning} -- a single episode 
in the curriculum streams all the data for the system to perform a single
task (e.g., multiclass classification).
\textit{(2) Task-Incremental Continual Learning} -- the system
is subjected to a sequence of
episodes, each comprising a specific task (e.g. two-class classification).
Each episode has a different
output head (e.g., two-class classifier).
Task labels are used both at
train and test time.
\textit{(3) Class-Incremental Continual Learning} --
episodes and curriculum are similar to 
task-incremental learning, except that the model learns
with a single (shared) output head. The final accuracy
measures the model's ability to predict on test data from
all the classes, spanning across episodes. The learning
algorithm utilizes the task labels only at train time. Thus,
this is much harder scenario and closer to ``task-free" continual learning.

\subsection{\label{sec:backprop} Transfer of bio-inspired design concepts to backPropagation-based algorithms}

A key challenge of bio-inspired models is that they are hard to
integrate with state-of-the-art approaches based on deep neural
networks. To overcome this problem and integrate the
core ideas of our model with other continual learning algorithms, 
we have transferred the key design concepts to a backprop setting. 
The key insight is that the MSE learning rule described above
closely resembles the change in weights expected from applying
an SGD optimizer with an MSE loss function.
Consequently, we can recreate the main features of our model
through a combination of optimizers, loss functions, custom
layers, and weight clamping. Our approach is summarized in Table \ref{tab:backprop}

From a network architecture perspective, the sparse layer is
computed directly using Eq. \ref{eq:dyn_thresh}, where the
mean and standard deviation are calculated sample by sample (i.e., in contrast to batch normalization approaches). 
The output of the sparse layer is then passed to
Eq. \ref{eq:F_l}, which is implemented as a linear layer with 
constant bias, whose output is passed through a
rectified linear unit and a hyperbolic tangent function.
We can view the synaptic plasticity rule as a combination of loss function
and optimizer. If we focus on the MSE rule, weight updates are given by:
\begin{equation}
    \label{eq:mse}
        \Delta W_l^{t+1} = l_r^t
        (\vect{x}_m^t-\vect{x}_o^t)\vect{x}_e^t 
\end{equation}
Eq. \ref{eq:mse} resembles the weight updates that would result from applying a mean square error loss
function followed by an SGD optimizer without momentum. Therefore, we have adapted
this combination in our backprop studies. This is in contrast to
most continual learning algorithms, which tend to focus on a cross-entropy loss function. Furthermore,
weight clamping is an essential feature of our bio-inspired model. Therefore, 
we have also introduced a clamping mechanism that can be invoked each time the optimizer is called.
Finally, the use of constant bias in the bio-inspired model coupled with a hard threshold leads to a high
probability of $\vect{x}_e$ and $\vect{x}_o$ being zero. This leads to modifications of the synaptic
weights as long as the output neurons are below that threshold. We can generalize this idea by introducing
modifications to the MSE rule that attempt to similarly minimize when learning takes place through the use of
a filtered MSE that returns a zero loss function whenever the maximum output corresponds to the target label.

\begin{table*}[t!]
\caption{Strategy to transfer the key innovations from the bio-inspired model into a BackPropagation-based Algorithm} 
\begin{center}
\begin{adjustbox}{max width=0.80\linewidth}
\begin{tabular}{|l|l|}
    \hline
    Bio-inspired design & Backprop-based algorithm \\
    \hline
    \hline
    Sparse projection layer & Sparse projection with fixed weights \\
    \hline
    Bounded spike rate & ReLU + Tanh / sigmoid activation functions \\
    \hline
    Fixed activation threshold & Constant bias + hard threshold \\
    \hline
    Local learning rule & Modified MSE loss function \\
                        & Weight clamping after optimization step \\    
    \hline
\end{tabular}
\end{adjustbox}
\end{center}
\label{tab:backprop}
\end{table*}

\section{Results and Discussion}

\subsection{Single-Task Online Learning}

To validate our architecture's ability to carry out online learning,
we tested it against the MNIST, Fashion MNIST \citep{xiao2017fashion} (F-MNIST), 
CIFAR-10, and CIFAR-100
datasets. These are prevalent in the literature on continuous learning. In all cases,
the networks were capable of achieving high accuracies in as few as $0.5$ epochs.
The comparison of the best accuracy obtained
for each learning rule among the configurations evaluated in the search for each dataset is
shown in Table~\ref{tab:combined2}a.
In Table~\ref{tab:combined2}b 
we show the classification accuracies for each of the five data sets obtained using the optimal configurations but run during the $1$ epoch.
We also compare our approach with the kernelized information bottleneck (KIB)~\cite{pogodin2020kernelized} approach, which is also an SGD-alternative local learning approach that is biologically plausible. Although we have a shallow network (approximately 100k learnable weights) and operate in online learning with $1$ training epoch, we obtain a test accuracy of $96.81$ on MNIST data, $85.22$ on F-MNIST data, and $73.24$ on CIFAR-10 data. These results are close to the accuracy of $98.1$ obtained by KIB (trained with 3 layers, each with 1,024 neurons of a fully connected network for 100 epochs) on the MNIST data. Our approach significantly outperforms KIB on the CIFAR-10 data. Our accuracies are also on par with (or outperform) other SGD-based shallow network architectures \citep{YanguasGil_MSE_2019,xiao2017_online,cohen2017emnist}.

\begin{table}[h!]
    \caption{(left) (a) Comparison of accuracy across datasets and learning rules; (right) (b) Classification accuracy in the single-task online learning scenario for MNIST, F-MNIST, and CIFAR-10 datasets using the optimal configurations learned through the optimization framework.}
    \label{tab:combined2}
    \begin{minipage}{.5\linewidth}
        \begin{center}
        \begin{adjustbox}{max width=0.8\textwidth}
        \begin{tabular}{|c|c|c|c|}
            \hline
        
             Rule/Data & MNIST & F-MNIST & CIFAR-10 \\
            \hline
            GEN & $31.20$ & $32.38$ & $32.39$    \\
            \hline
            OJA & $77.63$ & $63.18$ & $52.87$    \\
            \hline
            MSE & $96.16$ & $85.13$ & $77.37$  \\
            \hline
            INEL & $96.40$ &$85.09$& $77.96$   \\
            \hline
        \end{tabular}
        \end{adjustbox}
        \end{center}
    \end{minipage}%
    \begin{minipage}{.5\linewidth}
        \begin{center}
        \begin{adjustbox}{max width=\textwidth}
        \begin{tabular}{|c|c|c H|c|}
            \hline
            
             Algo/Data & MNIST & F-MNIST & E-MNIST & CIFAR-10\\
            \hline
            pHSIC (Shallow), 100 epoch & 98.1 &  88.8 & - & 46.4 \\
            \hline
             Ours, 1 epoch & 96.81 & 85.22 & 97.36 & 73.24\\
            \hline
        \end{tabular}
        \end{adjustbox}
        \end{center}
    \end{minipage} 
\end{table}

\subsection{Continual Learning}
We now discuss the results for the task-incremental and class-incremental continual learning scenarios with the Split-MNIST, Split-CIFAR-10, and  Split-CIFAR-100 continual learning benchmarks that have been extensively adopted in the literature \citep{shin2017continual,zenke2017continual,nguyen2017variational}. Split-MNIST data is prepared by splitting the original MNIST dataset (both training and testing splits) consisting of ten digits into five 2-class classification tasks. This defines a continual learning scenario in which the model sees these five tasks incrementally one after the other. The Split-CIFAR-10 is analogous to Split-MNIST, while for the split-CIFAR100 we follow~\cite{mai2021online,Lee2020A,chen2020mitigating} and construct 10 tasks each with 10 classes. Additional experimental details are discussed in Appendix~\ref{appendix:expt details}, and hyperparameter optimization for NNA is discussed in Appendix~\ref{NNA:optimization}.

\subsubsection{Task-Incremental Learning}

\begin{table*}[t!]
\caption{Classification accuracy for the task-incremental learning experiments on the Split-MNIST, Split CIFAR-10, and Split CIFAR-100 datasets. The accuracy metrics are reported as mean and standard deviation over $5$ repetitions.} 
\begin{center}
\begin{adjustbox}{max width=0.75\linewidth}
\begin{tabular}{|c|c|c|c|c|}
    \hline
          & Method & Split-MNIST & Split-CIFAR-10 & Split-CIFAR-100 \\ 
    \hline
    \hline
    \multirow{2}{*}{Baseline}& iid-offline  & $99.46 \pm 0.08$ & $95.51 \pm 0.22$  & $80.89 \pm 0.75$ \\
    & Fine-Tune  &  $97.29 \pm 0.96$ & $62.56 \pm 5.76$  & $47.54 \pm 1.61$ \\
    \hline
    \hline
    \multirow{3}{*}{\shortstack[l]{Continual \\ Learning \\ Memory-free }}& Online EWC  &$98.46 \pm 0.47$ & $59.98 \pm 2.27$  & $20.24 \pm 1.23$\\
     &SI&  $97.95 \pm 0.70$ & $65.71 \pm 1.79$  & $33.00 \pm 2.59$ \\
     &LwF&  $99.26 \pm 0.09$& $63.47 \pm 1.52$  & $19.45 \pm 1.19$  \\
     \cline{1-5}
     \multirow{5}{*}{\shortstack[l]{Continual \\ Learning \\ Memory-based \\ (Buffer=0.5k) }}&A-GEM  &  $99.16 \pm 0.08$ & $72.02 \pm 1.29$  & $38.39 \pm 1.98$ \\
     &iCaRL  &  $98.40 \pm 0.09$  & $82.01 \pm 0.76$  & $50.56 \pm 0.23$ \\
     &GSS  &  $97.80 \pm 0.80$  & $86.38 \pm 1.21$  & $56.86 \pm 1.68$ \\
     &RPSNet & -- & 67.0   & 40.1 \\

     &DER++  &  $99.29 \pm 0.02$ & $85.95 \pm 1.62$  & $58.28 \pm 1.50$ \\
     \cline{1-5}
     & {NNA-ST} (INEL) &  $65.13 \pm 0.66$  & $81.05 \pm 0.15$  & $23.04 \pm 0.26$ \\
     & {NNA-ST}  (MSE ) &  ${\bf 99.56} \pm 0.04$ & ${\bf 94.43} \pm 0.15$  & ${\bf 83.17} \pm 0.15$\\

    \hline
\end{tabular}
\end{adjustbox}
\end{center}
\label{tab:TaskINC-expts}
\end{table*}

To test the architecture's performance for task-incremental learning, 
we transferred the optimal configurations for the single-task learning case (\emph{NNA-ST})
to the Split-MNIST~\cite{farquhar2018towards}, Split-CIFAR-10, and Split-CIFAR-100 continual learning benchmarks~\cite{shin2017continual,zenke2017continual,nguyen2017variational}. For the specific case of MNIST, we also included the case of INEL and MSE to
evaluate the impact of the inelastic plasticity rule on the overall performance. We compared the accuracy of our approach with baseline, memory-free and memory buffer-based continual learning approaches, as well as with \textbf{ Fine tuning}, where the model is continuously trained upon arrival of new tasks without any strategies to avoid catastrophic forgetting. As baselines, we consider \textbf{iid offline}, where the model is evaluated in a noncontinual learning scenario and trained with multiple passes through the data, which are sampled iid.
Among the memory-free approaches, we choose \textbf{Online EWC}~\cite{schwarz2018progress}, \textbf{SI}~\cite{zenke2017continual},  and \textbf{LwF}~\cite{li2017learning}; and among the memory-based methods we compare with \textbf{A-GEM}~\cite{chaudhry2018efficient}, \textbf{iCaRL}~\cite{rebuffi2017icarl}, \textbf{GSS}~\cite{aljundi2019gradient}, \textbf{RPSNet}~\cite{NEURIPS2019_83da7c53}, 
, and \textbf{DER++}~\cite{buzzega2020dark} to cover the most recent works.  In all these models, we ensured that the hyperparameters used are consistent and report results averaged across 5 runs. 
For RPSNet, we report the 
metrics from the paper~\cite{chen2020mitigating}, since the experiments were run with the same 
hyperparameters as those used for other reported approaches in this work.
The comparison is shown in Table~\ref{tab:TaskINC-expts}, where we find that our approach obtains an accuracy of $99.56$, $94.43$, and $83.17$ on MNIST data,
Split-CIFAR-10, and Split-CIFAR-100, respectively. These experiments show that architectures based on local learning rules can outperform memory-free and memory-based approaches in the task incremental learning scenario, even though learning is confined to a single layer and without any explicit memory replay. Furthermore, we ran the Supermasks in Superposition model~\cite{wortsman2020supermasks} on Split-Cifar-100 (considered in their work) using hyperparameters from this work and found that its accuracy was 12.67, which is much lower than our approach. 
In addition, since the feature extraction layer of our NNA approach for the Split CIFAR-10 and Split CIFAR-100 experiments uses pretrained weights (see sec~\ref{feature extraction}), we evaluated the effect of the pretrained weights on other algorithms in Appendix~\ref{appendix:pretrain} and found that they generally do not provide a significant advantage over working directly with the inputs in the online Task-Incremental learning scenario considered in this work. However, a notable trend was that when adopting the sparse projection layer (with features derived from the pre-trained model) and a linear classification layer mirroring the methodology in our NNA approach, we observe an increase in the accuracy of memory-free approaches and A-GEM, but they still
fall short of the best accuracy obtained.

\begin{table*}[t!]
\caption{Classification accuracy for the class-incremental learning experiments on the Split-MNIST, Split CIFAR-10, and Split CIFAR-100 datasets. The accuracy metrics are reported as mean and standard deviation over $5$ repetitions}
\begin{center}
\begin{adjustbox}{max width=0.80\linewidth}
\begin{tabular}{|c|c|c|c|c|}
    \hline
          & Method & Split-MNIST & Split-CIFAR-10 & Split-CIFAR-100 \\ 
    \hline
    \hline
    \multirow{2}{*}{Baseline}& iid-offline  & $95.82 \pm 0.33$ &  $80.54 \pm 0.63$ & $48.092 \pm 0.90$\\
    & Fine-Tune  & $19.68 \pm 0.02$ & $19.19 \pm 0.06 $ &  $8.32 \pm 0.23$\\
    \hline
    \hline
    \multirow{2}{*}{\shortstack[l]{Continual \\ Learning \\ Memory-free }}& Online EWC  & $19.92 \pm 0.35$ & $16.18 \pm 1.37$ & $4.41 \pm 0.37$\\
     &SI&   $19.76 \pm 0.01$ & $17.27 \pm 0.87$ & $5.87 \pm 0.21$ \\
     &LwF&   $20.54 \pm 0.64$ & $18.53 \pm 0.1$2 & $6.93 \pm 0.32$\\

     \cline{1-5}
     \multirow{7}{*}{\shortstack[l]{Continual \\ Learning \\ Memory-based \\ (Buffer=0.5k) }}&A-GEM  & $48.57 \pm 5.26$ & $18.21 \pm 0.16$ & $6.18 \pm 0.20$\\
     &iCaRL   & $72.55 \pm 0.45$ & $35.88 \pm 1.43$ & $15.76 \pm 0.15$\\
     &GSS  & $54.14 \pm 4.68$ & $49.22 \pm 1.71$  & $11.33 \pm 0.40$\\

     & ER-MIR & $86.60 \pm 1.60$ & $37.80 \pm 1.80$ & $9.20 \pm 0.40$\\
     & CN-DPM & ${\bf 93.81}\pm 0.07$ & $47.05 \pm 0.62$ & $16.13 \pm 0.14$ \\

     &DER++  & $92.21 \pm 0.54$ & $52.01 \pm 3.06$ & $15.04 \pm 1.044$\\
     &ER-ACE & $82.98 \pm 1.79 $ & $35.16 \pm 1.34$ & $8.92 \pm 0.25$ \\

     \cline{1-5}

     & {NNA-CIL} ({\small INEL+MNIST}) & $77.25 \pm 1.02$ & $45.95 \pm 0.90$ & ${\bf 25.56} \pm 0.69$ \\
     & {NNA-CIL} ({\small INEL+CIFAR10}) & $60.82 \pm 2.00$ & ${\bf 52.55} \pm 2.05$ & $18.87 \pm 0.33$ \\

    \hline
\end{tabular}
\end{adjustbox}
\end{center}
\label{tab:TaskINC-offline-expand}
\end{table*}

\subsubsection{Class-Incremental Learning}

We followed the same approach
described above for the more
challenging class-incremental learning scenario
on the Split-MNIST, Split-CIFAR-10, and Split-CIFAR-100 datasets. 
We employ the configuration obtained by explicitly optimizing the learning rule configuration for
class-incremental learning (\emph{NNA-CIL}).
In Table~\ref{tab:TaskINC-offline-expand}, we compare the accuracy of our approach in the context of class-incremental learning with baseline, memory-free, and memory buffer-based continual learning approaches. We use the same baselines and memory-free approaches as those adopted in task-incremental case; but for the memory buffer-based, we also compare against \textbf{ER-MIR}~\cite{aljundi2019online}, \textbf{CN-DPM}~\cite{Lee2020A} and \textbf{ER-ACE}~\cite{caccianew} 
, which are the most recently proposed approaches.
With \emph{NNA-CIL} we see a significant increase in performance (with both configurations obtained by optimizing on Split-MNIST and Split-CIFAR-10 datasets, respectively) on all three datasets. The accuracies obtained on all three datasets are significantly (in some cases, 3X) higher than the accuracies of the memory-free algorithms. In the case of split-CIFAR-10, the resulting accuracy of $52.55$ with \emph{NNA-CIL} is also on a par with $52.01$ obtained with the memory-based DER++ algorithm, while for split-CIFAR-100, \emph{NNA-CIL}'s accuracy of $25.56$ significantly outperforms that of other approaches. In addition, we ran the Supermasks in Superposition model~\cite{wortsman2020supermasks} on Split-MNIST (considered in their work) using hyperparameters from this work and found that its accuracy was 40.87, which is much lower than that of our approach. 
Similarly to the task-incremental learning scenario, we evaluated the effect of the pretrained weights on other algorithms for the Split CIFAR-10 and Split CIFAR-100 experiments in the class-incremental learning scenario (see Appendix~\ref{appendix:pretrain}). We found that even for this case, the pretrained weights generally decrease the accuracy of the considered continual learning algorithms over working directly with the inputs in the online Class-Incremental learning scenario. One notable exception is that of the iCaRL applied to Cifar-10 
but with Resnet-18 model trained using features extracted from the pretrained model, where we observe an improvement in accuracy over training with original data but still falls short of the best accuracy.

\begin{figure}[t!]
\centering
  \includegraphics[width=0.5\linewidth]{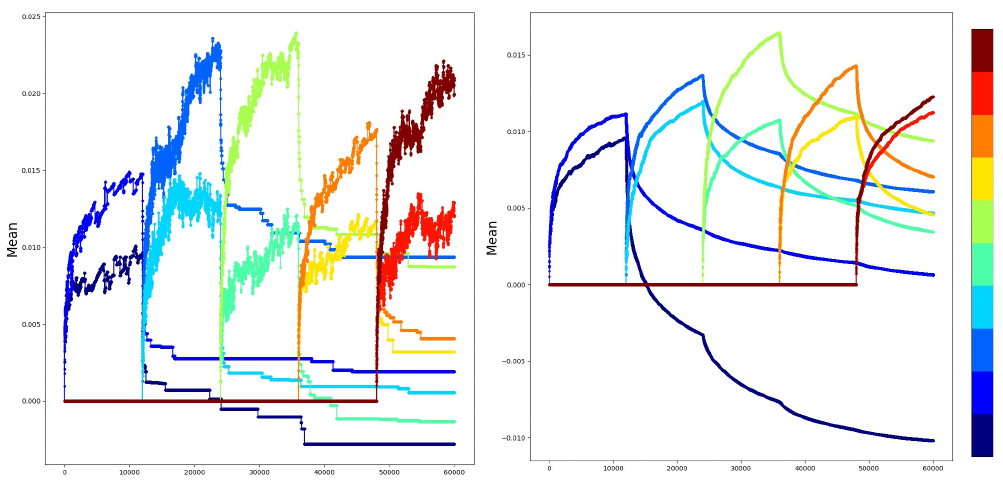}
  \includegraphics[width=0.48\linewidth]{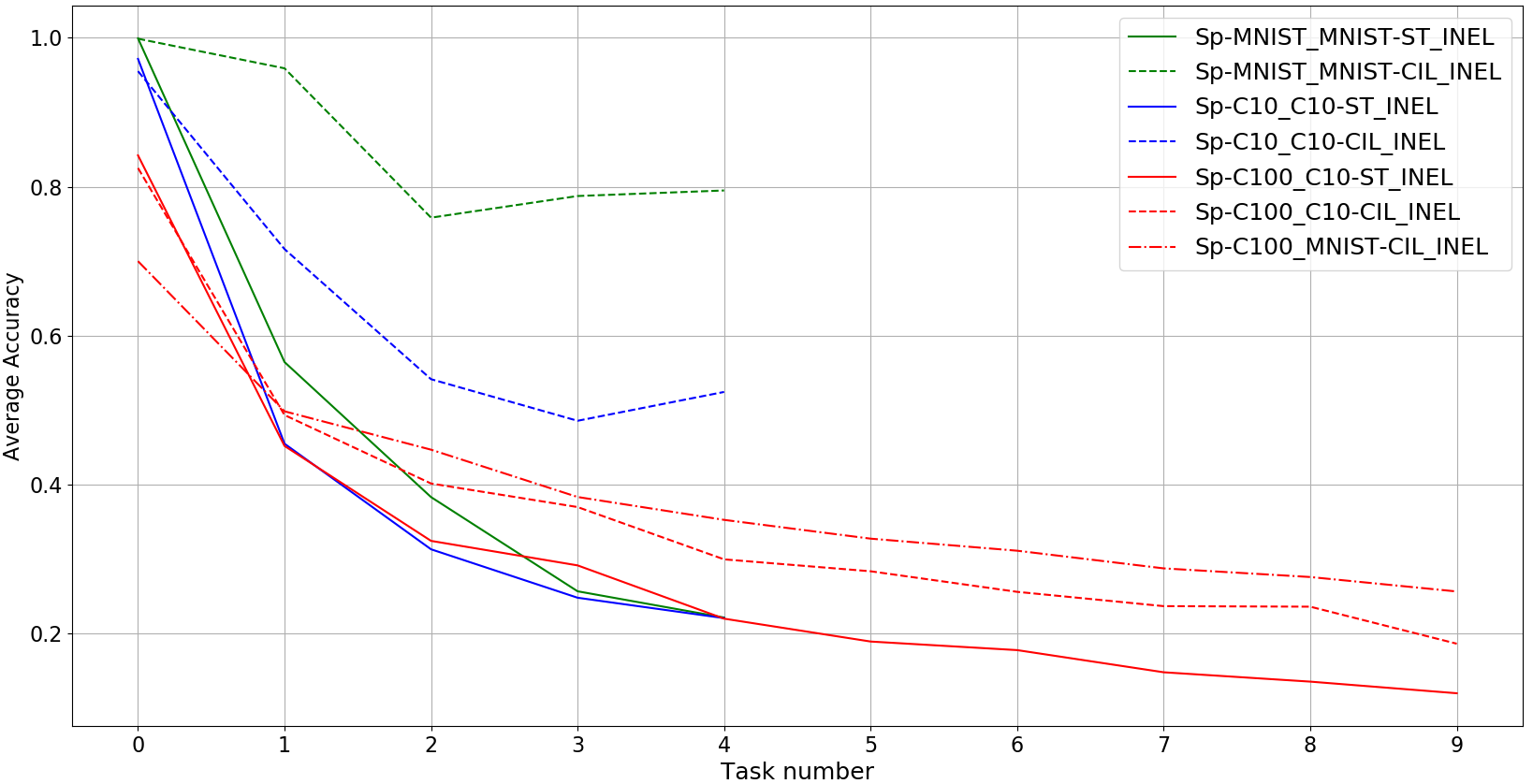}
\caption{Comparison of the evolution of $W_l^t$ for split MNIST learned in the class-incremental learning scenario using (a) \emph{NNA-ST}+INEL configuration shown on left-most plot and (b) that  \emph{NNA-CIL}+INEL shown in middle plot. The right-most plot shows the average accuracy as a function of task for each of the datasets (Split-MNIST, Split-CIFAR10) run with the hyperparameter configuration obtained from NNA-ST and NNA-CIL with INEL rule for their respective data. For Split-CIFAR100, we compare with INEL configurations from Split-CIFAR-10 and NNA-CIL+INEL from Split-MNIST. We see that NNA-CIL configurations consistently have higher average accuracy.}
\label{fig:SMNIST_Cinc}
\end{figure}
The higher performance of the modulated inelastic rule can be attributed to the stabilization of the most significant weights for each class. This rule provides relative stabilization and is sensitive to changes in other synapses reaching the same neurons, where the sensitivity is modulated by hyperparameters. Consequently, since weight consolidation is not critical for the single-task learning, the optimal configurations do not fully exploit this feature, leading to threshold values that are not compatible with those of class-incremental experiments, which require stabilization over longer periods of time. The comparison of weight evolution of class-incremental learning on Split-MNIST data with \emph{NNA-ST} and \emph{NNA-CIL} (Figure~\ref{fig:SMNIST_Cinc}) further highlights that the design principles necessary for single-task might differ from class-incremental learning. Further discussion can be found in the Appendix~\ref{sec: appendix1}.

\begin{table*}[h!]
\scriptsize
\caption{Accuracy due to transfer metalearning from single task online learning to task-incremental and and class-incremental continual learning experiments.} 
\begin{center}
\begin{tabular}{|c|c|c H |c|c|c|c|c|c|}
 \hline
      & \multicolumn{4}{c|}{Single Task Learning (0.5 epoch)} & &\multicolumn{2}{c|}{Task-Incremental Learning} &\multicolumn{2}{c|}{Class-Incremental Learning}\\
    \hline
      Config ($\downarrow$), Dataset ($\rightarrow$)  & MNIST  & F-MNIST  & E-MNIST & CIFAR10 & & SplitMNIST & SplitCIFAR-10 & SplitMNIST & SplitCIFAR-10\\
    \hline 
    \hline
      MNIST (w/INEL) & $96.39$  & $85.48$ & -- & $72.91$ & & $65.25$ & $77.63$ & $21.56$ & $21.19$\\
    \hline
      MNIST (w/MSE) & $96.16$   & $80.75$  & -- & $69.87$ & & $99.60$ & $92.50$ & $21.84$ & $20.33$\\
    \hline
      F-MNIST & $94.52$  & $85.13$  & -- & $71.29$ & & $99.25$ & $94.37$ & $21.39$ & $21.19$\\
    \hline
      CIFAR-10 (w/INEL)& $93.99$   & $84.13$  & -- & $77.95$ & & $64.99$ & $81.20$ & $20.86$ & $22.33$\\
    \hline
      CIFAR-10 (w/MSE) & $94.45$  & $84.21$  & -- & $77.37$ & & $99.03$ & $94.65$ & $21.75$ & $22.85$\\
    \hline
\end{tabular}
\end{center}
\label{tab:Transfer-datasets}
\end{table*}

\subsection{Transferability Study} 
\vspace{-0.1in}
Identification of the best configuration for each group of classes (a dataset) 
is based on the assumption that all the data are available at the beginning of 
the training procedure. However, for many online learning
scenarios, both continual and single-task learning, this 
configuration might not be known a priori. This raises the question of transfer metalearning: \textit{how to
effectively optimize a learning
algorithm to learn unknown tasks and data}.
To address this problem, we empirically
studied the effect of transferring optimal configurations
to the continual learning scenario,
and across datasets for the single-task learning case. 

\begin{wrapfigure}[17]{r}{0.56\textwidth}
\centering
  \includegraphics[width = 0.83\linewidth]{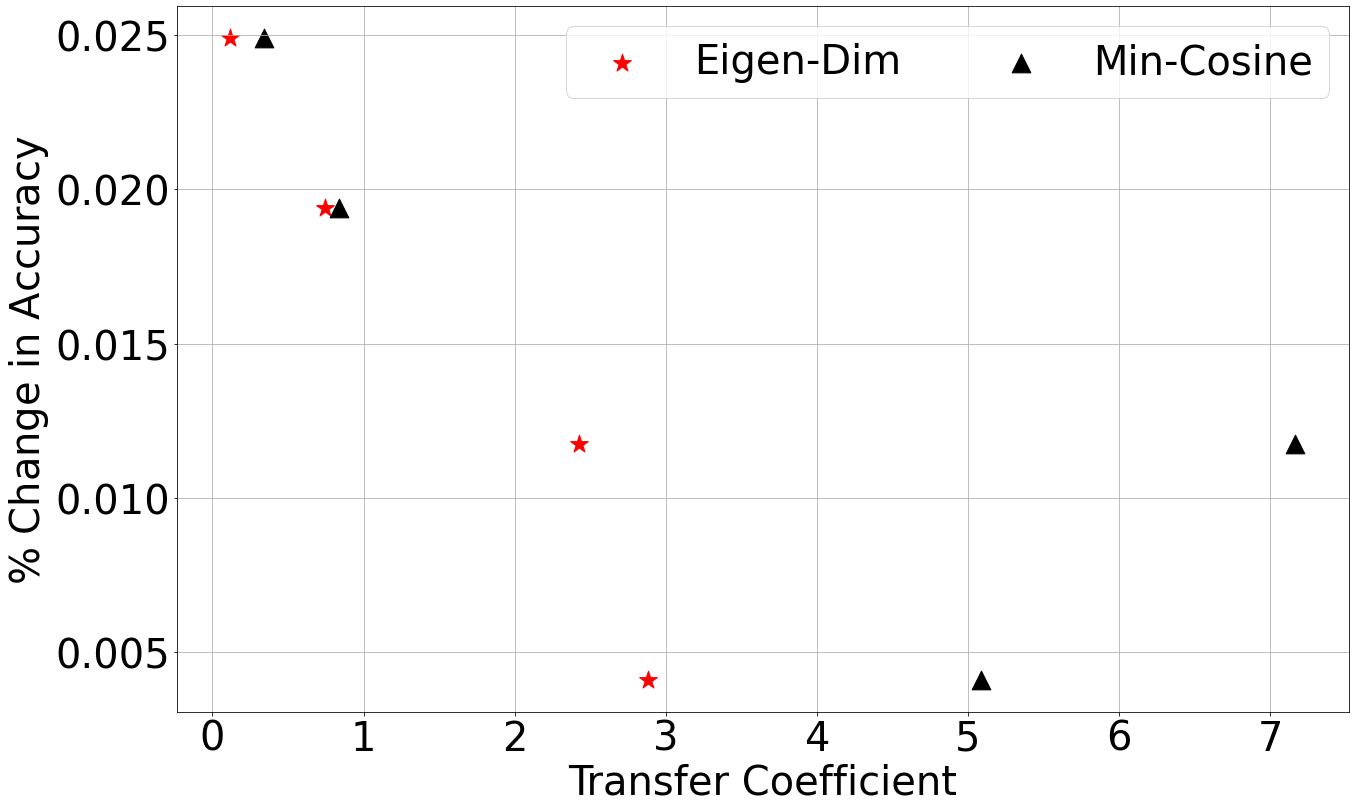}
\caption{\small Drop in classification accuracy  during  hyperparameter configuration transfer as a function of a transfer coefficient obtained for eigen dimension and minimum cosine distance metrics.}%
\label{fig:trnsfr}
\end{wrapfigure}
In Table~\ref{tab:Transfer-datasets} we show the accuracy resulting from transferring
optimal conditions across tasks and across datasets. From these results we can
extract several conclusions. First, the transfer from a single task online to task-incremental
continual learning is excellent for the MSE rule but not for the inelastic rule. We
attribute this effect to the strong role that the synaptic weight distribution plays in
the inelastic rule. Second, for the class-incremental case, the accuracies obtained
when transferring hyperparameters are significantly lower than those achieved when
conditions are specifically optimized for class-incremental learning. This result
indicates that, at least in memory-constrained situations, learning rules must
be highly optimized for class-incremental problems and that we must rely on surrogate data to carry out transfer metalearning. 

\noindent{\bf Distance metrics as a measure of transferability:}
To rationalize the results shown in Table \ref{tab:Transfer-datasets}, we have explored the correlation
between performance drop across datasets and their
effective dimensionality and the distance between datasets.
{\it (a) Eigen dimension}: We consider as a metric of
dimensionality the sum of squares of the eigenvalues ($\lambda$) of the covariance matrix of the complete dataset, so that
    $d = (\sum_i\lambda_i)^2/\sum_i \lambda_i^2$.
This metric
has been used in the past to characterize sparse representations \cite{Litwin_2017}.
{\it(b) Cosine distance}: We calculate the distance between two datasets using the minimum cosine distance between any two categories (see details in Appendix~\ref{appendix:transfer}).
Figure \ref{fig:trnsfr} shows the
drop in classification accuracy during transfer metalearning
as a function of a \emph{transfer coefficient}
obtained from two different metrics: the relative
difference in the dimension of the eigenvalue and the minimum
cosine distance between the two datasets. Both values
are normalized as
$D = {|{M_1-M_2}|}/{M_1}$,
where $M_1$ is the metric of the dataset
selected to run the experiment and $M_2$ is the metric
of the dataset whose optimal configuration is used. Note that this
transfer coefficient is not symmetric because of the different
normalization, as should be expected, since transfer
metalearning is directional. The results clearly show that as the distance between the datasets increases, transfer learning the configurations across them leads to a decrease in accuracy.

\section{Application to backpropagation-based approaches}

In order to evaluate whether the main contributions of bio-inspired networks can be applied to existing
backpropagation-based continual learning algorithms, we ported some of the key innovations into a backpropagation-based
model described in Section \ref{sec:backprop} and implemented in Avalanche~\cite{lomonaco2021avalanche}. This allowed us to integrate the networks,
optimizers, and loss functions into other existing strategies. In particular, we chose to integrate
our model with the Naive and ER-ACE algorithms, which are the worst and best performing approaches across
the memory-free and memory-based approaches for the class-incremental learning setting.
We found that the accuracy of online Naive (Fine-tune) improved from $19.68$ to $41.10$ and ER-ACE improved
from $82.98$ to $86.06$ for the class-incremental SplitMNIST benchmark when we introduced the bio-inspired modifications
summarized in Table \ref{tab:backprop}.
These preliminary results show that the biologically-inspired design patterns are not only effective at mitigating catastrophic forgetting through local learning rules, but also have the potential to improve the backpropagation-based continual learning approaches in the literature. This result is a step forward towards combining the strengths of biological mechanisms with modern deep learning to obtain more accurate and robust continual learning approaches.

\section{Summary and Conclusions}

In this work, we have applied bio-inspired design principles to the problem of memory-free online continual learning. Our architecture relies on local learning rules to carry out single task online and continual learning tasks. 
We demonstrate that our approach is able to obtain accuracies on par with other SGD-alternative approaches in an online learning scenario.
In the task-incremental continual learning scenario, we obtained accuracies exceeding 99\%, 94\%, and 83\% respectively for
Split-MNIST, Split-CIFAR-10, and CIFAR-100, thus exceeding the state-of-the-art of
memory-based and memory-free algorithms. 
In class-incremental learning,
our architecture clearly outperforms memory-free algorithms in the Split-MNIST, Split-CIFAR-10, and Split-CIFAR-100 tests. It achieves an accuracy of 52.55 and 25.56 for Split-CIFAR-10 and Split-CIFAR-100, respectively, that also exceeds that of memory-based algorithms. 

An important fraction of the highest-performing configurations identified in this work relied on a novel inelastic rule that implements a simple form of memory consolidation for synaptic weights that deviate from the pool of presynaptic weights of each neuron. This rule leads to the long-term stabilization of weights that are relevant for a specific class, mitigates catastrophic forgetting. This mechanism can be easily combined with any synaptic plasticity mechanism to stabilize networks learning from non-stationary data. Configurations using this rule achieved accuracies that are three times higher than the best results obtained by using memory-free algorithms in the class-incremental tasks considered in this work. This result highlights the potential of moving beyond conventional local learning rules to fully realize the possibilities afforded by recurrent neural networks to control learning and plasticity in feedforward neural networks.

A key characteristic of our model is that we provide depth to the learning layer through a sparse projection layer that highlights the most relevant feature combinations. This allows us to sidestep the problem of credit assignment when using local learning rules. On the other hand, this also provides a limitation of the proposed approach, since the plasticity for supervised learning is restricted to a single layer in our architecture.
In conclusion, our NNA approach's salient design principles include the biologically inspired neural architecture and loss formulation, effectively mitigate catastrophic forgetting. We have demonstrated that these principles can be integrated into existing continual learning algorithms, leading to improved online continual learning. These findings have significant implications for developing intelligent systems that can learn continuously from incoming data streams without catastrophic forgetting.

\subsubsection*{Acknowledgments}
Primary development of this work was funded by the DARPA Lifelong Learning Machines (L2M) Program. This work has also been partially supported by the DOE Office of Science, Advanced Scientific Computing Research, and SciDAC programs under Contract No.\ DE-AC02-06CH11357. The computational resources of the Argonne Leadership Computing Facility, which is a DOE Office of Science User Facility supported under the DE-AC02-06CH11357 contract, and the Laboratory Computing Resource Center (LCRC) at the Argonne National Laboratory have been utilized for this work.

\newpage
\appendix

\section{Appendix}
\subsection{Impact of configuration on catastrophic forgetting}
\label{sec: appendix1}

To better understand why the configurations learned in the single-task scenario transferred well to the
Task-Incremental scenario but not to the Class-Incremental case, we studied
the evolution with time of the weight matrix ($W_l$) in the
neuromodulated learning layer,
which has dimensions of $W_l$ of $x_e \times x_o$.
More specifically, we tracked the evolution
of the mean weight presynaptic to each
output neuron: by averaging $W_l$ across $x_e$, we produce a $x_o$
dimensional array for every data sample streamed to the model during learning. 
We use this to track learning
for each class. 
The evolution of this
mean weight is shown in Figure~\ref{fig:SMNIST_Cinc} (a)
for the Split-MNIST data in the class-incremental learning scenario.
The left plot corresponds to the
INEL configuration transferred from
the single task learning, while the
bottom plot represents the
evolution of the mean synaptic weight for the configuration learned by optimizing over the class-incremental learning itself. Each
task is composed of 12k samples.

After a specific task is concluded,
the mean synaptic weights of the classes
learned during that task suffer a
significant drop, which is consistent
with the presence of catastrophic forgetting
and lower accuracies. However, the magnitude
of this drop is significantly
more gradual in the configuration
optimized for the continual learning
case. This correlates with
the good class incremental learning accuracy observed with this configuration. This
more gradual decay is a consequence of
the inelasticity of the learning rate, which remain zero for weights that differ
significantly from the mean in the
INEL rule.
To evaluate the transferability of the
configuration obtained from
optimizing class-incremental learning
across datasets, we applied this
very same configuration, learned
on Split-MNIST, to the Split-CIFAR-10
and Split-CIFAR-100 datasets.
The resulting accuracies,
$45.95$ for Split-CIFAR-10 and $25.56$
for Split-CIFAR-100,  outperform most
of the memory-free and
memory-based methods we compared (Table \ref{tab:TaskINC-offline-expand}).

Furthermore, we calculate the average forgetting metric (in addition to 
average accuracy shown in Figure~\ref{fig:SMNIST_Cinc}) as a function of task id. Average Forgetting at task $\mathcal{T}_i$ is defined as the average loss in accuracy on a task $\mathcal{T}_j$ after training on a sequence of tasks $\mathcal{T}_1 ... \mathcal{T}_i$. Let the test accuracy on task $\mathcal{T}_j$ after learning with task $\mathcal{T}_i$ be $a_{i,j}$, then the average forgetting can be defined as~\cite{mai2021online}:

\begin{ceqn}
\begin{gather} 
\text{Average Forgetting} (F_{i})=\frac{1}{i-1} \sum_{j=1}^{i-1} f_{i, j} \nonumber\\
\text{where~} f_{k, j}=\max_{l \in\{1, \cdots, k-1\}}(a_{l, j})-a_{k, j}, \forall j<k
\label{eq: fgt}
\end{gather}
\end{ceqn}

\begin{figure}[h!]
\centering
  \includegraphics[width=0.6\linewidth]{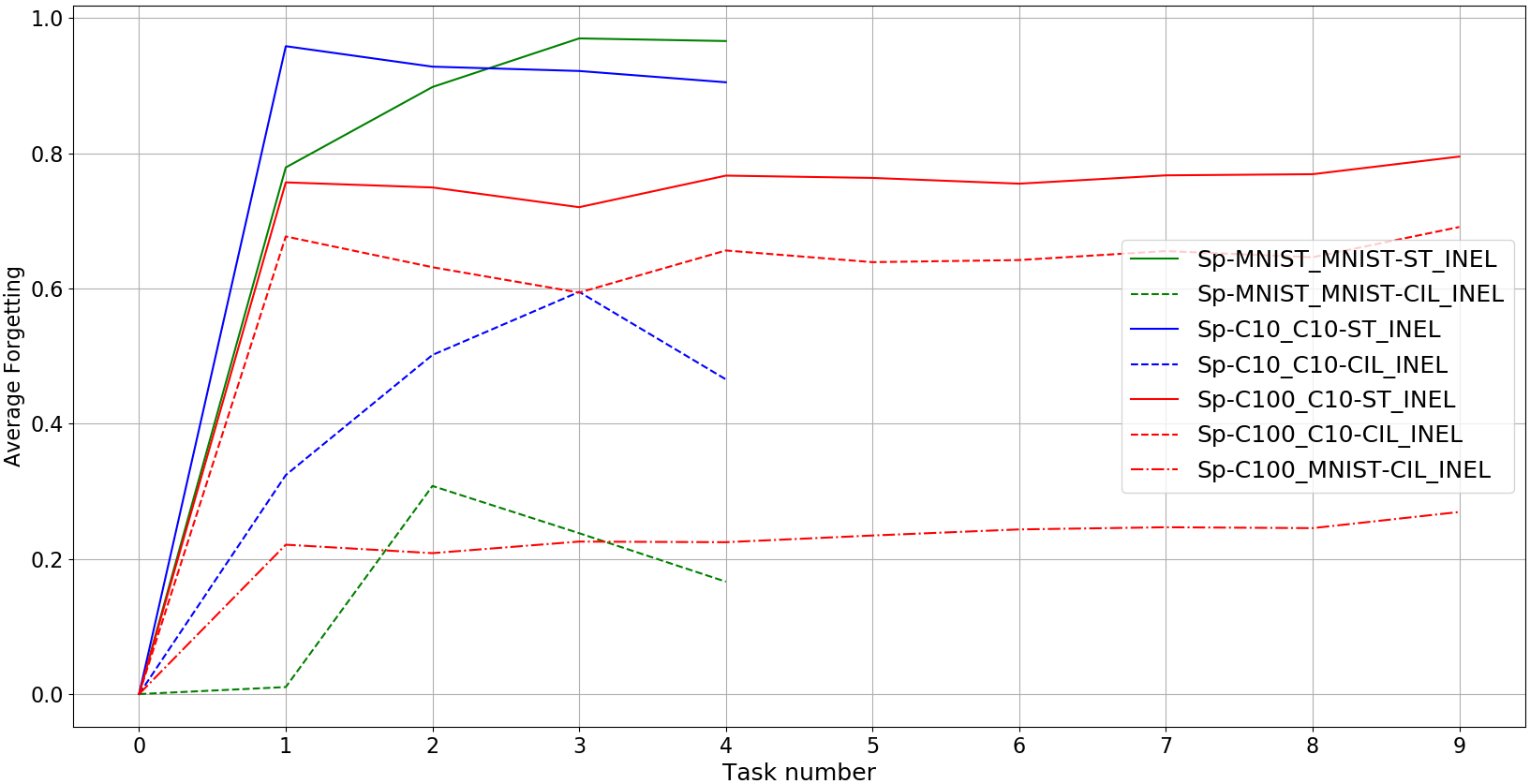}
\caption{. 
Average forgetting as a function of task for each of the datasets (Split-MNIST, Split-CIFAR10) run with the hyperparameter configuration obtained from NNA-ST and NNA-CIL with INEL rule for their respective data. For Split-CIFAR100, we compare with INEL configurations from Split-CIFAR-10 and NNA-CIL+INEL from Split-MNIST. We see that NNA-CIL configurations consistently have lower average forgetting.
}
\label{fig:avg_forgetting}
\end{figure}

    The $F_i$ values for Split-MNIST, Split-CIFAR10, Split-CIFAR100 trained in the class-incremental learning scenario are shown in Figure~\ref{fig:avg_forgetting}. We find that the NNA-CIL configuration
consistently gives the lowest average forgetting, which was also giving the highest average accuracy.

\subsection{Effect of pretrained weights on the continual learning performance}
\label{appendix:pretrain}
In this section, we evaluate the impact of using pretrained weights adopted in our Neuromodulated Neural Architecture (NNA) on the other continual learning algorithms evaluated for the CIFAR-10 and CIFAR-100 datasets. We considered the Online EWC, SI, LWF, which are memory-free approaches and the A-GEM, iCarl, GSS, and DER$++$ which are the memory-based approaches considered in the previous experiments and presented in the main text. We measure the performance of these continual learning algorithms by calculating the classification accuracy in the following scenarios: (1) Utilize the sparse projection layer and a linear classification layer to assess the classification accuracy based on features derived from the pre-trained model, mirroring the methodology in our NNA approach; (2) Bypass the use of original inputs and instead feed the Resnet-18 model with features extracted from the pre-trained model; (3) Replace the Resnet-18 model with a two-layer Multi-Layer Perceptron (MLP) for processing the pre-trained features.

We evaluate the continual learning performance for these three scenarios in both the task-incremental and the class-incremental learning scenarios and the results are shown in Table~\ref{tab:Pretrain_Cifar10} for CIFAR-10 and Table~\ref{tab:Pretrain_Cifar100} for CIFAR-100. We performed a grid search to find the hyperparameters that perform the best for each of the algorithms in the task-incremental and class-incremental learning settings. The hyperparameters considered in the grid search for these algorithms are the same as those presented in the work~\cite{buzzega2020dark}, where the range of each parameter is discretized into four values and an exhaustive search is performed to find the best configuration. Finally, we report the accuracy on the test data averaged across five runs.

From these results in Table~\ref{tab:Pretrain_Cifar10},\ref{tab:Pretrain_Cifar100} we observe that the pre-trained weights in general do not appear to provide a significant advantage over working directly with the inputs in the online continual learning scenario considered in this work. However, we do see some trends, first one is that specifically in Scenario 1, the memory-free algorithms and A-GEM show increase in the accuracy when using the pre-trained weights for task-incremental learning setting, but they still fall short of the best accuracy obtained on both the CIFAR-10 and CIFAR-100 datasets. Another observed trend is that Scenario 2 provides the highest accuracy for the memory-based approaches in the task-incremental learning setting for both the datasets; however, they also do not surpass the accuracy obtained without pre-trained weights. Finally, we see that the class-incremental learning accuracy consistently decreases across all scenarios considered with pre-trained weights for both datasets except for iCaRL applied to Cifar-10 with scenario 2 where we see significant accuracy gain over No pretrain case but is still falls short of the best accuracy obtained across algorithms in the no pretrain case.

\begin{table*}[t!]
\caption{Classification accuracy for the task-incremental and class-incremental learning experiments on the Split CIFAR-10 dataset in three scenarios that characterize the impact of using pre-trained weights on continual learning with the baseline approaches. Accuracy metrics are reported as mean and standard deviation on $5$ repetitions.}
\begin{center}
\begin{adjustbox}{max width=\linewidth}
\begin{tabular}{|c|c|c|c|c|c|c|c|c|c|}
 \hline
      &  &\multicolumn{4}{c|}{Task-Incremental Learning} &\multicolumn{4}{c|}{Class-Incremental Learning}\\
    \hline
          & Method & NoPretrain & Scenario 1 & Scenario 2 & Scenario 3 & NoPretrain & Scenario 1 & Scenario 2 & Scenario 3\\ 
    \hline
    \hline
    \multirow{3}{*}{\shortstack[l]{Continual \\ Learning \\ Memory-free }}& Online EWC & $59.98 \pm 2.27$ & $81.73 \pm 0.96$ & $57.36 \pm 3.29 $  & $68.42 \pm 2.83$ & $16.18 \pm 1.37$ & $16.61 \pm 0.34$ & $15.68 \pm 1.65$ & $16.56 \pm 0.25$\\
     &SI& $65.71 \pm 1.79$ & $81.92 \pm 0.43$ & $64.00 \pm 4.17$  & $67.81 \pm 1.43$ & $17.27 \pm 0.87$ & $16.88 \pm 0.37$ & $17.70 \pm 0.32$ & $16.82 \pm 0.13$\\
     &LwF& $63.47 \pm 1.52$ & $76.82 \pm 0.37$ &  $61.75 \pm 1.80$ & $68.10 \pm 1.19$ & $18.53 \pm 0.12$  & $16.67 \pm 0.54$ & $18.39 \pm 0.41$ & $16.51 \pm 0.42$\\
    \hline
    \hline
     \multirow{4}{*}{\shortstack[l]{Continual \\ Learning \\ Memory-based \\ (Buffer=0.5k) }}&A-GEM  & $72.02 \pm 1.29$ & $81.49 \pm 0.59$ & $71.77 \pm 3.32$  & $73.72 \pm 1.03$ & $18.21 \pm 0.16$ & $17.16 \pm 0.29 $ & $18.29 \pm 0.63$ & $16.88 \pm 0.07$\\
     &iCaRL  & $82.01 \pm 0.76$ & $69.64 \pm 0.12$ & $87.84 \pm 0.88$  & $79.16 \pm 0.53$ & $35.88 \pm 1.43$ & $28.62 \pm 0.12$ & $46.57 \pm 2.54$ & $36.63 \pm 0.28$\\
     &GSS  & $86.38 \pm 1.21$ & $77.38 \pm 0.56$ & $83.75 \pm 1.45$  & $76.95 \pm 0.68$ & $49.22 \pm 1.71$ & $29.35 \pm 1.20$ & $26.28 \pm 1.91$ & $26.46 \pm 0.64$\\
     &DER++  & $85.95 \pm 1.62$ & $81.03 \pm 0.23$ & $86.66 \pm 0.85$  & $77.80 \pm 0.14$ & $52.01 \pm 3.06$ & $34.64 \pm 0.73$ & $41.36 \pm 3.22$ & $26.76 \pm 0.68$\\
     \cline{1-8}

    \hline
\end{tabular}
\end{adjustbox}
\end{center}
\label{tab:Pretrain_Cifar10}
\end{table*}

\begin{table*}[t!]
\caption{Classification accuracy for the task-incremental and class-incremental learning experiments on the Split CIFAR-100 dataset in three scenarios that characterize the impact of using pre-trained weights on continual learning with the baseline approaches. Accuracy metrics are reported as mean and standard deviation on $5$ repetitions.}
\begin{center}
\begin{adjustbox}{max width=\linewidth}
\begin{tabular}{|c|c|c|c|c|c|c|c|c|c|}
 \hline
      &  &\multicolumn{4}{c|}{Task-Incremental Learning} &\multicolumn{4}{c|}{Class-Incremental Learning}\\
    \hline
          & Method & NoPretrain & Scenario 1 & Scenario 2 & Scenario 3 & NoPretrain & Scenario 1 & Scenario 2 & Scenario 3\\ 
    \hline
    \hline
    \multirow{3}{*}{\shortstack[l]{Continual \\ Learning \\ Memory-free }}& Online EWC & $20.24 \pm 1.23$ & $49.60 \pm 0.33$ & $29.54 \pm 3.97$ & $23.60 \pm 1.43$& $4.41 \pm 0.37$ & $5.22 \pm 0.07$& $6.86 \pm 0.30$ & $4.95 \pm 0.13$\\
     &SI& $33.00 \pm 2.59$ & $50.04 \pm 0.26$ & $28.33 \pm 3.49$ & $22.28 \pm 0.59$ & $5.87 \pm 0.21$ & $5.25 \pm 0.11 $ & $5.39 \pm 0.49$ & $4.50 \pm 0.19$ \\
     &LwF& $19.45 \pm 1.19$ & $47.01 \pm 0.34 $ & $22.36 \pm 1.56$ & $27.80 \pm 0.59$& $6.93 \pm 0.32$  & $5.43 \pm 0.17$& $6.87 \pm 0.37$ & $5.01 \pm 0.31$\\
    \hline
    \hline
     \multirow{4}{*}{\shortstack[l]{Continual \\ Learning \\ Memory-based \\ (Buffer=0.5k) }}&A-GEM  & $38.39 \pm 1.98$ & $48.18 \pm 0.27$ & $39.79 \pm 2.10$ & $23.93 \pm 1.42$& $6.18 \pm 0.20$ & $5.1 \pm 0.15$ & $7.10 \pm 0.19$ & $4.93 \pm 0.16$\\
     &iCaRL  & $50.56 \pm 0.23$ & $34.57 \pm 0.16$ & $46.70 \pm 0.42$ & $38.56 \pm 0.52$ & $15.76 \pm 0.15$ &$10.04 \pm 0.22$ & $10.72 \pm 0.49$ & $9.78 \pm 0.24$\\
     &GSS  & $56.86 \pm 1.68$ & $38.17 \pm 0.56$ & $54.27 \pm 1.15$ & $36.22 \pm 1.00$& $11.33 \pm 1.40$ & $6.88 \pm 0.36$ & $8.57 \pm 0.42$ & $6.91 \pm 0.19$\\
     &DER++  & $58.28 \pm 1.50$ & $46.59 \pm 0.43$ & $56.71 \pm 1.24$& $33.73 \pm 0.53$& $15.04 \pm 1.04$ & $8.24 \pm 0.40$ & $10.01 \pm 0.79$ & $7.90 \pm 0.60$\\
     \cline{1-8}

    \hline
\end{tabular}
\end{adjustbox}
\end{center}
\label{tab:Pretrain_Cifar100}
\end{table*}

\subsection{Additional Experiment Details}
\label{appendix:expt details}

The experiments for Split MNIST use a multilayered perceptron model
with two layers and $400$ nodes per layer, while the Split CIFAR-10 and Split CIFAR-100 experiments use a ResNet-18 model.
The experiments for Split MNIST, Split CIFAR-10 and Split CIFAR-100 datasets using the models 
iid-offline, Fine-Tune, Online EWC, SI, LwF, A-GEM, iCaRL, GSS, DER++ has been run by adopting
the code in \url{https://github.com/aimagelab/mammoth}. Experiments with ER-MIR have been run 
by adopting the code in \url{https://github.com/optimass/Maximally_Interfered_Retrieval}. The 
experiments with CN-DPM adopted the implementation in \url{https://github.com/soochan-lee/CN-DPM}.
The experiments with OnlineER-ACE adopted the implementation in \url{https://github.com/ContinualAI/avalanche/}. In all the above models, we ensured that the hyperparameters used are consistent with the
online continual learning scenario mentioned above. 

\subsection{Architecture Configuration Optimization}
\label{NNA:optimization}

\begin{figure}[h!]
\centering
\includegraphics[width=0.7\textwidth]{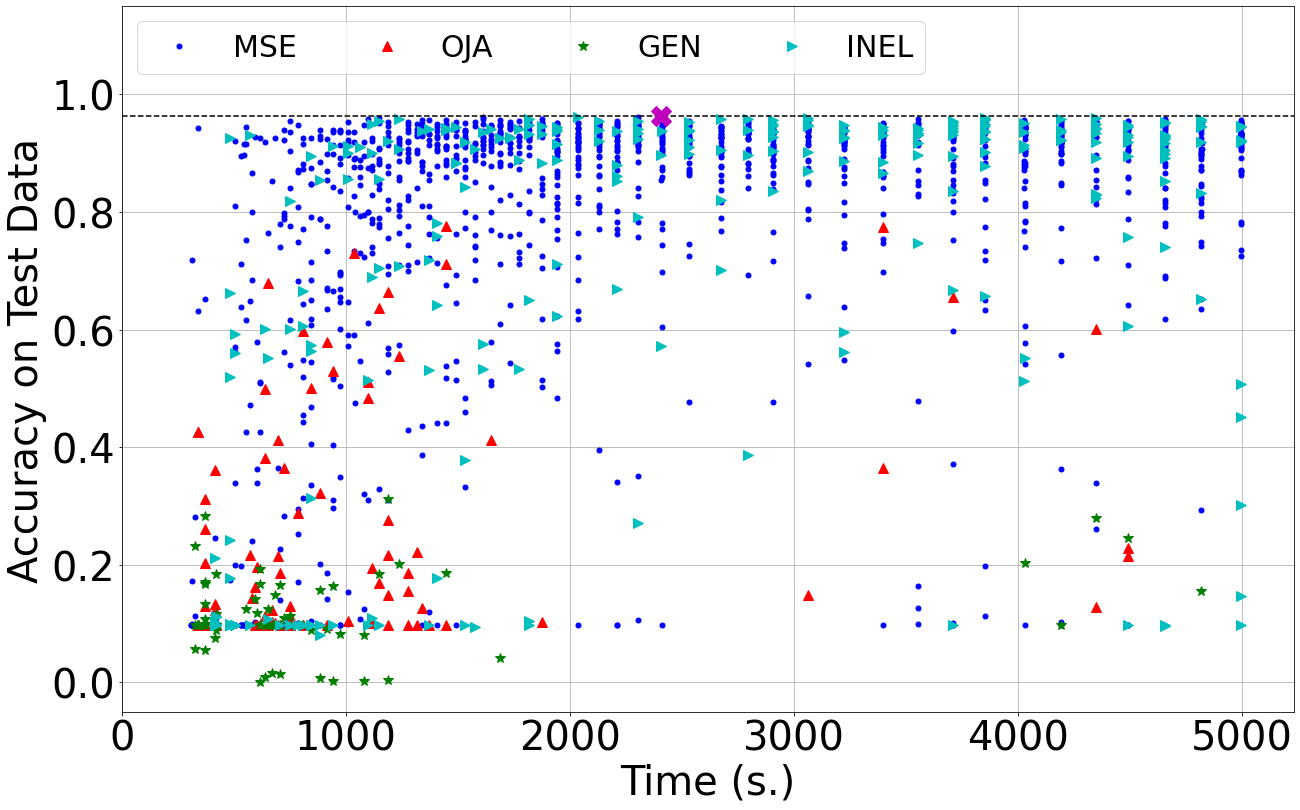}
        \caption{Search trajectory obtained for MNIST dataset from the mixed-integer black-box optimization. The configurations are colored by their evaluated learning rule, with blue, red, green and cyan corresponding to \MSE., \OJA., \GEN., and \INEL. respectively. The accuracy on test data is plotted as a function of wallclock time, which highlights the diversity in the configurations evaluated as the search progressed}
        \label{fig:search_traj}
\end{figure}

The parameter space in the proposed multilayer neuromodulated architecture is composed of categorical variables (e.g., the selection of the local learning rule), integer parameters (e.g., the dimension of the hidden layer) and continuous parameters in each of the learning rules. We adopt Bayesian optimization~\cite{Balaprakash_DH_2018} to find the high-performing parameter configuration in this mixed search space (categorical, integer, continuous). We used a holdout (validation) set of 10k samples split from the training data to perform the hyperparameter optimization.
An example search trajectory showing the evolution of the final accuracy as the algorithm searches multiple configurations for MNIST is shown in \ref{fig:search_traj} . The data, color coded according to the selected learning rule,
shows how initially the algorithm searches across the different learning rules, until
it quickly settles on the most promising rule. The comparison of the best accuracy obtained
for each learning rule among the configurations evaluated in the search for each dataset
shown in Table~\ref{tab:combined2}.

\subsection{Transferability Study}
\label{appendix:transfer}
The Eigen dimension and cosine distance between the datasets used to generate Figure~\ref{fig:trnsfr} are shown in Tables~\ref{tab:pcadim} and Table~\ref{tab:distances} respectively.
\begin{table}[ht!]
\caption{Eigen dimension metric for each of the datasets.}
\begin{center}
\begin{adjustbox}{max width=0.6\textwidth}
\begin{tabular}{|c|c|c|c| c|}
    \hline
      Data  & MNIST      & F-MNIST    & E-MNIST    & CIFAR-10 \\
    \hline 
      Eigen-Dim   & $30.69$ & $7.91$ & $27.09$ & $132.60$ \\
    \hline
\end{tabular}
\end{adjustbox}
\end{center}
\label{tab:pcadim}
\end{table}
\begin{table}
\centering
\caption{Minimum and maximum cosine distances
    between centroids of the categories of different datasets}
\begin{adjustbox}{max width=0.6\textwidth}
\begin{tabular}{|c|c|c|c| H|}

    \hline
      Data  & MNIST      & F-MNIST    & E-MNIST  \\
    \hline 
      MNIST & $(0.073,0.55)$ & $(0.17,0.54)$ & $(0.044,0.59)$ \\
    \hline
      F-MNIST &          & $(0.012,0.56)$ & $(0.13,0.57)$ \\
    \hline
      E-MNIST &        &          &   $(0.098,0.52)$  \\
    \hline
     
\end{tabular}
\end{adjustbox}
\label{tab:distances}
\end{table}
Another recent work~\cite{NEURIPS2020_f52a7b26} characterized the distances between datasets of same dimensionality using the geometric dataset distances obtained from optimal transport (OTDD). We observe that the minimum cosine distance presented in Table~\ref{tab:distances} follows the same trend as the pair-wise distances between MNIST, F-MNIST, and E-MNIST obtained from OTDD.
Since the cosine similarity considered in this work is restricted to the datasets with same sized inputs, we consider the Singular Vector Canonical Correlation Analysis (SVCCA)~\cite{raghu2017svcca} metric to characterize the distance between inputs of different sizes. The SVCCA metric combines singular value
decomposition and canonical correlation analysis, wherein first a singular value decomposition is applied to a dataset to obtain the directions to explain $99\%$ of variance and then a pairwise canonical correlation is calculated between the data that is intended to be compared. Specifically, we calculate this metric between each of the classes in the dataset pairs as done with the cosine distance. However, we use the sum of the off-diagonal elements in the distance matrix normalized with the number of elements for the same datasets and all the elements for distance matrix for dissimilar datasets. The values obtained are shown in Table~\ref{tab:distances_svcca}, but we found that it might not be a reliable metric, as all the distance values are nearly the same, with variation only in the last few decimal places. We will explore other similarity metrics in future work.

\begin{table}[h!]
\centering
\caption{The SVCCA metric between datasets}
\begin{adjustbox}{max width=0.6\textwidth}
\begin{tabular}{|c|c|c|c|c|}

    \hline
      Data  & MNIST      & F-MNIST    & E-MNIST & CIFAR-10  \\
    \hline 
      MNIST & $0.0488$ & $0.0491$ & $0.0484$ & $0.0531$\\
    \hline
      F-MNIST &          & $0.0483$ & $ 0.0486$ & $0.0529$ \\
    \hline
      E-MNIST &        &          &   $0.0492$ &  $0.0533$\\
    \hline
      CIFAR-10 &        &          &           & $0.0534$ \\
    \hline
\end{tabular}
\end{adjustbox}
\label{tab:distances_svcca}
\end{table}

\subsection{Ablation Studies}
We study the sensitivity of the results to the various components of the proposed 
Neuromodulated Neuromorphic Architecture through a systematic ablation study and 
illustrate the results using the MNIST dataset.  We first consider the single-task 
learning scenario with the MNIST-ST (INEL) configuration that corresponds to the
accuracy reported in Table~\ref{tab:combined2}a  
for these experiments. In the first experiment, the neuromodulated learning layer
is replaced by a multilayer perceptron model with two layers, each with 400 nodes, 
while keeping the feature extraction layer intact. In this scenario, we found
a drop in test accuracy to 37$\%$ (with 10 epochs) from the $96.84\%$ obtained with
the proposed architecture. 
Next, we consider both feature extraction and neuromodulated learning
layers, but turn off the sparsity imposed by the activity sparsity 
(Equ.~\ref{eq:dyn_thresh}). This lead to an accuracy drop to $23.24\%$,
which signifies the importance of dynamic thresholding. Finally, we removed
the feature extraction layer and directly fed the images to the
neuromodulated learning layer. In this case, we obtain an accuracy of
$90.72\%$, which underlines the importance of the feature extraction layer
(sparse projection + activity sparsity in this work) in improving the accuracy.

\end{document}